# Child Drawing Development Optimization Algorithm based on Child's Cognitive Development

Sabat Abdulhameed[1] and Tarik A. Rashid[1,*]
[1]Computer Science and Engineering Department, University of Kurdistan Hewler, Erbil, KRG, Iraq.
*Corresponding email: tarik.ahmed@ukh.edu.krd

**Abstract**

This paper proposes a novel metaheuristic Child Drawing Development Optimization (CDDO) algorithm inspired by the child's learning behaviour and cognitive development using the golden ratio to optimize the beauty behind their art. The golden ratio was first introduced by the famous mathematician Fibonacci. The ratio of two consecutive numbers in the Fibonacci sequence is similar, and it is called the golden ratio, which is prevalent in nature, art, architecture, and design. CDDO uses golden ratio and mimics cognitive learning and child's drawing development stages starting from the scribbling stage to the advanced pattern-based stage. Hand pressure width, length and golden ratio of the child's drawing are tuned to attain better results. This helps children with evolving, improving their intelligence and collectively achieving shared goals. CDDO shows superior performance in finding the global optimum solution for the optimization problems tested by 19 benchmark functions. Its results are evaluated against more than one state of art algorithms such as PSO, DE, WOA, GSA, and FEP. The performance of the CDDO is assessed, and the test result shows that CDDO is relatively competitive through scoring 2.8 ranks. This displays that the CDDO is outstandingly robust in exploring a new solution. Also, it reveals the competency of the algorithm to evade local minima as it covers promising regions extensively within the design space and exploits the best solution.

## 1. Introduction

Metaheuristic algorithms have been being used to solve today's real-world optimization problems, which can not be solved by traditional mathematical techniques [1]. They are defined as higher-level strategies guiding the heuristic procedures that are more problem-specific to increase their performance [2]. They are gaining more popularity by solving real-world optimization problems that cannot be solved within a time limit and using a specific or deterministic method [3], [4]. Metaheuristic optimization techniques are used to solve complex problems in industry and services areas like engineering, management, and medical health [5]. Nature-inspired metaheuristics are of the most popular classes of such techniques [6], [1]. Some of the algorithms such as the Firefly algorithm and Particle swarm optimization have been implemented broadly and frequently in several applications due to their high efficiency [7]. While some other algorithms are well established to solve specific, critical and challenging problems, such as protein structure prediction and the travelling salesman problem.

There are several comparative advantages of the metaheuristics such as being stochastic, as they start with random solutions. At the same time, the derivation of the search space is not required, which makes it capable of solving problems with unknown derivatives. Another advantage is that metaheuristics guide the search process to avoid falling into local optima due to the strategies that are more based on randomness [8], [9]. Although there is massive work in this area of research, it remains to be one of the hottest topics in Artificial Intelligence as it is being logically proved by the theorem of "no free lunch" that there no widespread set of rules, which could tackle all optimization problems. It also demonstrates that there is still a need for work in this area since most algorithms may have good results on some problems while performing poorly in solving other problems [10]. Improvement in both computation complexity and faster convergence rate are among the areas of work in this field of research to obtain better results to solve optimization problems.

Despite the outstanding results achieved by applying nature-inspired algorithms to find the optimal solutions, there are still some considerable gaps between the theory and practicality of these algorithms. One of the main gaps are the shortcomings in the mathematical analysis in the following aspects; solutions accuracy balance and computational efforts, as well as parameter tuning and control [11]. Considering the gaps and shortcomings of the existing algorithms motivates researchers to develop new ones that are novel with practical techniques to overcome the current challenges in solving is recommended [12]. New algorithms that have a stable convergence rate, a solid framework to control and tune their parameters, and have a good performance with a reasonable time limit are required [5], [13].

The CDDO algorithm is motivated by the child's learning behaviour and cognitive development in the early childhood years. The use golden ratio is a ratio commonly found in nature, and when used in art, it increases organic and natural-looking compositions that are visually pleasing to the eye. CDDO uses golden ratio and mimics cognitive learning and child's drawing stages using the parameters such as hand pressure width, length and golden ratio of the child's drawing. Through trial and error, the children's drawings evolve from scribble to an art piece.

The contribution of this study is the novelty of the notion behind the suggested algorithm that implements the method of child's cognitive development and child's drawing development stages backed up by its solid mathematical base using the golden ratio. Both techniques have never been used before in any metaheuristic researches, neither in computer science nor in nature-inspired algorithms. On the other hand, this algorithm contributes to an upcoming area of socio-inspired metaheuristics, which is a new approach that seeks inspiration from human behaviour [14].

The clear and straightforward layout of this research overcomes the significant challenges in the research area regarding the theoretical background. Unlike other metaheuristic algorithms, this study has a precise mathematical and theoretical base, such as the usage of the golden ratio method.

The sections of this paper outlined as follows; Section 2 includes the literature review, which provides evaluation and assessment of previous works done in optimization algorithms, particularly metaheuristic algorithms. Section 3 describes the basic theoretical background of a child's artistic development in early childhood. Later the relation between a child's artworks and their cognitive development is examined. The new algorithm in section 4 will be enlightened in detail with the mathematical model and used equations. Then, the logical flow of the algorithm was demonstrated. Section 5 contains the testing process along with the benchmark functions explanation. Next, it displays the results of the tested functions comparing with the metaheuristic and state of the art algorithms. Section 6 sums up the central answers to this research and advises its future direction.

## 2. Literature Review

Metaheuristics are relatively simple as they are inspired mainly by natural concepts. The inspirations are typically related to physical phenomena, animals' behaviours, or evolutionary concepts. The simplicity of these ideas makes it easy for scientists to apply and stimulate the biological concepts to develop, improve or hybridize two or more metaheuristics algorithms. Besides the simplicity, CDDO uses the golden ratio to find new solutions, avoid local optima and converge toward better results which is a new approach used in this field of study. Flexibility is another reason why metaheuristics are popular. It is the applicability of these algorithms to solve different problems without making any significant changes in the structure of the algorithm that makes them flexible. CDDO has a straightforward layout that can be easily applied after providing the input parameters. Another significant characteristic of metaheuristics is the process of solving any problem doesn't depend on the derivation of the input, which makes it highly satisfactory when applied to real problems with impracticable or unknown derivative information. Finally, meta-heuristics have superior abilities to avoid local optima compared to conventional optimization techniques [8].

Swarm Intelligence (SI) is mainly inspired by nature, particularly, biological systems that the agents follow straightforward rules while not having a centralized control structure-directing how each agent should behave in a relatively random process locally and led to intelligent behaviour on a global scale that is not intended by individual agents [15]. Genetic Algorithm (GA) is a crucial part of evolutionary computation. John Holland developed GA in mid the 1960s and became popular in the 1970s. It is a computational technique inspired by a biological process from Darwin's theory of evolution. GA is a metaheuristic approach, which emulates typical selection practice. The size of the population, mutation rate, crossover rate, and several generations are the main four Performance factors of GA [16]. In 1992, Ant Colony Optimization (ACO) was invented by Dorigo as part of his doctoral proposal, which is inspired by ants and the way they are searching for food, which is an elementary rule [17]. The ants go out from their colony and randomly go in different directions, and at any time, if one of them found a food source, then it yields to the colony, and on the way back, it leaves a trail of pheromone, which is a chemical substance. Zong Woo Geemet al. [18] in 2001 established the Harmony Search approach (HS). HS is a metaheuristic too. Unlike other metaheuristic algorithms. HS is not based on swarm intelligence; it is inspired by music. It is stated that musical harmony can serve as a good model for inventing a new technique. ABC was introduced by Karaboga in 2005. It is amongst the standard algorithms, and it is based on bee colony optimization. ABC was established based on the method used by real bees for food searching and communicating information about the food sources among the bees in the area. There is a difference between ABC and real Bee Colony Optimization since only scout and onlooker bees are used in equal size while setting up the initial population[19]. In 2008, Xin-She Yang presented the Firefly algorithm (FA). The inspiration for this approach was taken from the tropical behaviour of fireflies. FA is one of the algorisms that can be easily implemented at the same time it is flexible. There is a vast number of firefly species, and they produce rhythmic flashes; for some of the particular species, the flashes are unique [20]. In 2009, another optimization technique, which is called Cuckoo Search or CS, was established by Xin-she Yang and Suash Deb. It is enthused by some cuckoo species' behaviour called brood parasitism [12].
Bat Algorithm (BA) is another metaheuristic approach established by Xin-She Yang in 2010 [21], [22]. The inspiration was taken from microbats and their behaviour of echolocation. They can engender great echolocations, which are natural means used by bats to detect an object by utilizing the reflected sound from the object. By using this mechanism, bats can distinguish between the prey and the obstacles in the environment.

Some of the recent studies done in this area are the donkey and smuggler optimization algorithm or DSO developed by Shamsaldin, A.S., Rashid, T.A. in 2019. There are two main modes, the Smuggler and Donkeys, that are established to implement the searching and route selection behaviour in DSO. The smuggler mode is responsible for discovering the potential shortest paths. While in Donkeys mode, more behaviours are utilized, such as; running, facing & suiciding, and Facing & Supporting [23]. Fitness Dependent Optimizer (FDO) is another algorithm inspired by the process of bees' reproductive swarming, where typically, scout bees explore new nests. In principle and technically, FDO has no links with the ABC algorithm regardless of their inspiration source. The search agents of FDO are guided toward optimality using the weights generated by the employed fitness function [24]. Another new meta-heuristic method called Arithmetic Optimization Algorithm (AOA) is also an algorithm recently developed that utilizes the distribution behaviour of the leading arithmetic operators in mathematics, including (Multiplication, Division, Subtraction, and Addition). Almost similar approaches were used in both AOA and CDDO, which is the mathematical concept used in formulating the algorithm [25].

## 3   Algorithm Theory and Child's Cognitive Development

The brain can learn a pattern of repetitive events that strategy has been used in Artificial Intelligence researches and, more specifically, in machine learning. In a neural network, any unit establishes an output in the form of simple numerical functions of the received input. Any entity is represented by patterns that are distributed among the brain cells. One method of learning is learning by imitation. For example, babies learn languages by imitating what adults are saying. Another learning strategy is learning by analogy, which includes observing similarities between situations or between problems. Later the child applies the primarily learned metaphors to similar situations and experiences. Machine learning uses that mechanism of causal knowledge, which relies on the abilities of machines to make fundamental descriptions for any occurrences based on specific learning samples [26].

On the other hand, it has been demonstrated in studies about child cognitive development that babies can differentiate simple visual forms from birth, such as a cross versus a circle. In addition to that, babies can categorize what they see and form a general representation. Then, they compare them together and distinguish the features that can co-occur together. Another significant finding is that children at very young ages can learn characteristics of different objects, and they compare and interrelate between various features. The exact statistical learning mechanism is used by babies' brains in dynamic displays and learning the torsional changes in the process of analyzing which objects or events follow each other [27]. According to studies in children psychology's research, there are strong relations among drawings of children and their cognitive, social, and emotional expansion. It is also stated that the child's drawing can indicate the level of the children's cognitive development. The following subsections outline the primary inspiration sources of the CDDO algorithm, which are the golden ratio and child's drawing development stages.

### 3.1  Golden Ratio

In art, there is a big question about assessing any art piece since there are many factors to consider when it comes to labelling any art piece as beautiful or creative. By nature, any talented human being who practised art techniques, in some ways, knows how to create the most appealing drawings, such as Leonardo da Vinci's human body sketch in Figure 1. This may not hold true for beginners, particularly children, but improvement can be noticed through continuous practising. This phenomenon is like some kind of calculation to assess the proportion of drawing [28].

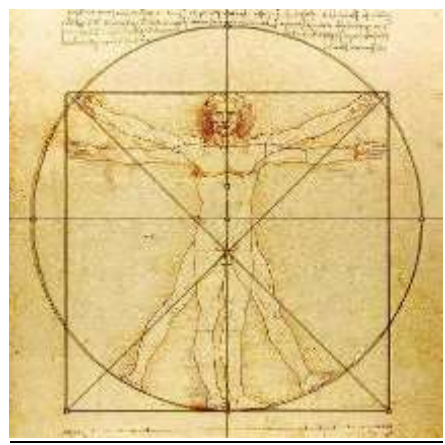

**Figure 1: The Human Body by Leonardo da Vinci** [29]

This makes it very easy for children or any talented artist to draw pleasing drawings most of the time without knowing anything about the calculation for finding perfect angles and proportions. Scientists have confirmed a strong relation between eye-pleasing contents with a mathematical ratio called Golden Ratio that is signified as φ, which is an illogical number= 1.6180339887. Due to its unique properties, it has been used in many fields, such as architecture, art, and design, applied in outstanding works of aircraft, sculptures, paintings, and architecture. The golden ratio can be found in the natural world in many forms, such as the body proportion of living beings, insects, and growth patterns of plants [30]. Mathematical series and geometrical patterns are used to represent the Golden Section (GS). In theory, the golden section is a point where any line segment can be divided into two sections keeping a unique property such that the ratio between the larger and smaller section will be equal to the ratio between the actual line and the larger section. As it is shown in Figure 2, point B is the line of AC is the golden section where the line AC can be divided into two sections, p and q, such that (AB=p and BC=q). According to the golden section theory, the ratio of the newly generated line sections p and q is equal to the ratio between (p + q) and p under the condition of p>q, which the relationship can be presented in the following equation.

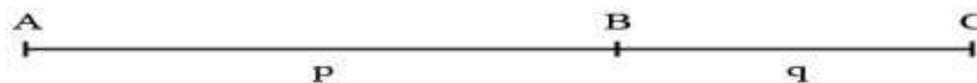

**Figure 2:A-line segment AC is divided in the golden section point B**

$$\varphi = +\frac{p}{q} + \frac{(p+q)}{p} \cong 1.6180339887 \qquad (1)$$

The relation demonstrated in the golden section equation shows that any section of the line AC is considered as 1; the remaining section can be found using Phi (φ). For instance, if q =1, that implies p= φ. As it is claimed from ancient times, the golden section is the optimum pleasing point at which the line can be sectioned. Later this idea was inherited in most of the ideal and flawless artworks, architectural design, and mathematical analysis.

### 3.2 Drawing Development Stages

Children's drawing is evolved through certain stages related to chronological age, which includes 6 to 18 years old. The stages of development have been studied in the early 1900s. The results showed that generally, children progress through certain stages of development in their art-making, and that is recognized through the survey [26]. In standard, these stages have some specific characteristics that appear in the respected stage according to the pattern that appears in their artwork. It is easier to control the tools like crayons than paint and a brush. There are a classification of age and characteristics due to the existence of particular traits that are repeatedly appear to occur in the child's drawing, which has been put into stages according to chronological age (particularly from 18 months to 6 years). Multiple internal and external factors directly have an impact on the development of a child's artistic abilities. For that, researchers in this area have suggested several theoretical models after working on them over the years. In this research, a specific model by [31] has been considered, which is representing the behaviours of a child in the process of drawing development. The five stages of the model being outlined in the following subsections that include detailed features in each stage.

#### 3.3.1   First stage: The Scribbling Stage (ages 2-4)

In this stage, young children are figuring out their hand's movement results in the strokes and writings they would perceive on the page. The random movements are either linear or curvature since the child is observing the movements, as the example is shown in Figure 3. Throughout time, the child's tolerance for the sensory input will increase, and later any other activity can be added into their routine.

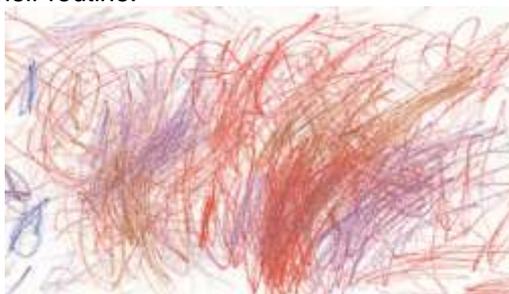
**Figure 3:Random scribble** [32]

#### 3.3.2   Second stage: The Pre-schematic Stage (ages 4-7 pattern memory)

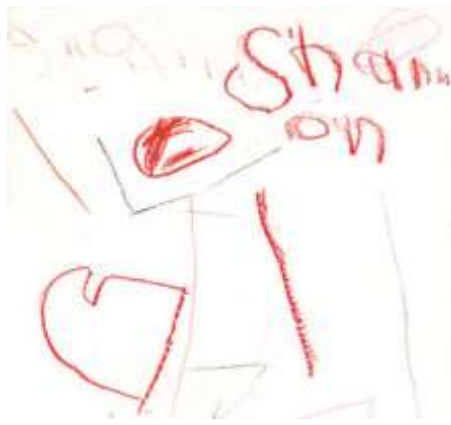

**Figure 4: Controlled scribbles** [32]

Throughout this stage, the child's drawings are more standard and copied versions of other drawings from the surrounding environment. This process comes from the child's ability to develop their drawing by practice and use the feedback of previous moves to recreate new scribbles. Another characteristic of this stage is searching for new ideas while having a basic schema in mind to draw elementary things. By this time, other individual patterns will begin to appear, and series of these patterns that are called "schema" will evolve in the child's approach and will be used as an immediate representation of events and places [33]. The following are the main points in the second stage, forming the mathematical model.

1. Controlled scribbling: it is when the child is starting to have control of the scribbling in terms of hand pressure shown in Figure 4.
2. Learning shapes (enhancement): Child's improvement in the shapes in size and perspective compared to the real ratio of the object of interest shown in Figure 5.
3. Patterns come by learning to control the movement and direction to create shapes.

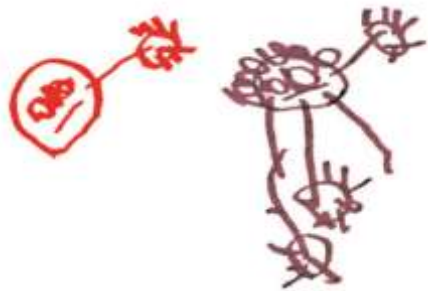

**Figure 5: Head and Feet Symbols** [32]

### 3.3.3 Third stage: The Schematic Stage (ages 7-9)

In this stage, the child will reach a level to draw a schema of the surrounding objects. The concept of schema is when the child gets to repeat the same pattern as it is observed in Figures 6-7. Children can apply the skills learned, observe patterns from real pictures, and try to give meanings to drawings, and practice continues to improve their drawings[33].

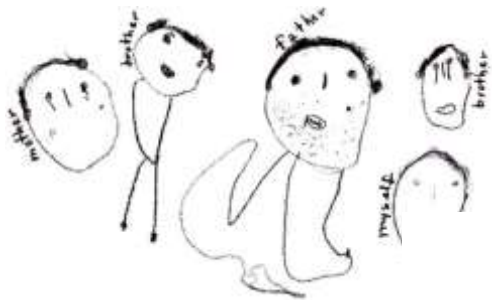

**Figure 6 A family portrait** [32]

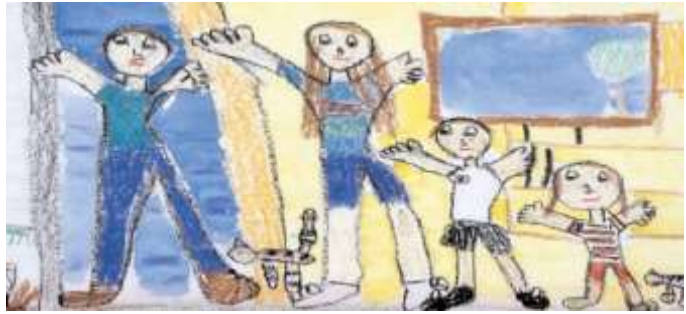

**Figure 7:A family portrait** [32]

### 3.3.4   Fourth stage: The Dawning Realism Stage (ages 9-11)

At this age, creating groups or ganging up with other children is referred to as social independence from adult interventions since they are trying to learn social structure in their personalized approaches. Children are getting more critical of themselves and their surroundings in addition to paying attention to detail and making an effort to draw realistic and naturalistic. The children will utilize and combine previous skills to come up with new ideas, which show their creativity in creating unique and more visually appealing drawings [33]. The following are the main points figured out in the fourth stage that has a direct connection with the mathematical model.

1) Object and people: appearing in the drawing to be realistic according to many factors, including the relation between the objects, distance, and perspective.
2) Imagination (creativity)

### 3.3.5   Fifth stage: The Pseudo-Naturalistic (Stage ages 11-13)

Awareness in this stage is higher than ever before though they are not yet an adult. Adding more detail and stepping toward self-centred approaches for drawing is noticed in this stage, such as the example shown in Figure 8.

.

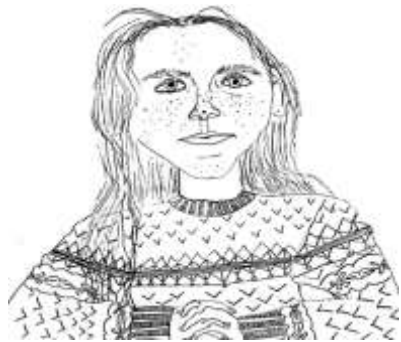

**Figure 8: A portrait of a classmate by a twelve-year-old** [32]

The main considered behaviours in this algorithm are aforementioned in the preceding sections, each playing a role in updating solutions converging toward an optimum solution. The main behaviours can be summarized as following; pattern memory, golden ratio consideration, properties of the drawing (hand pressure, drawing's length and width), coping from each other [33].

## 4   Mathematical Model of CDDO Algorithm

Artist's works seek the most appealing state that is in one way can be determined by the golden ratio appears in the artwork, as the optimization approaches pursue the finest situation, which is a global optimum that can be decided through evaluation function. Each drawing is determined by a set of factors the artist implies while drawing. In this case, the elements that have a role in a child's drawing throughout a child's artistic development stages were utilized to form a metaheuristic algorithm. The child's behaviour and the solid mathematical foundation of the golden ratio introduced in the previous section are used as an inspiration source for developing the mathematical equations for the new Human Behaviour inspired algorithm [34]. In the upcoming sub-sections, the mathematical equations of pattern memory, golden ratio, properties of the drawing (hand pressure, drawing's length and width), and cooperation between group members outlined with a detailed explanation of each section accordingly.

## *4.1 CDDO Stages*

A detailed description of each stage is explained as follows:

### 4.1.1 The first stage (The Scribble)

The first scribbles of a child's trials to discover drawing are predominantly random marks. The child in this stage is observing and discovering movement and hand pressure. Movement can be both linear and curvature randomly since the child is observing liner movements result in the lines and any other movements of hand results in curves. In this stage, the hand pressure is not reasonable; either too high or too low later will be improved by trial in the upcoming stages, taking into account many other factors—*initialization Xij for i=1 to N solution.*
$X$ is the current solution representing a child's drawing that has different decision variables such as hand pressure, golden ratio, length, and width of the drawing when several decision variables are noted as *i* and the number of parameters reported as *j*.

### 4.1.2 The second stage (Exploitation)

During this stage, the child learns to create shapes by controlling the movement and direction. At this stage, the drawings are more standard and copied, and the child is comparing his/her drawing to the best pattern he/she learned and defines the best sketched drawing so far, also of recreating new scribbles by imitating the best surrounding artist and comparing their drawing with the best sketched drawing so far by the group.

Hand pressure is one of the factors to classify a child's performance. While the hand pressure is relevant, the child's level is high. In the meantime, that indicates a child has enough skills to sketch a drawing with low hand pressure and an accurate golden ratio. At first, a Random Hand Pressure (*RHP*) is generated using equation 2. *RHP* is a random number between the lower boundary of the problem (LB) and the upper boundary of the solution (UP), which is a factor generated to evaluate the current solution's hand pressure with a current solution's hand pressure (*HP*). *HP* will be chosen among the parameters of the current solution by using equation (3) when *HP* is hand pressure, and *j* are several solution parameters.

$$RHP = rand(LB, UP) \qquad (2)$$
$$HP = X(i, rand(j)) \qquad (3)$$

### 4.1.3 Third stage (Golden ratio)

The child is now in the stage of applying the skills that have been learned from experiences and uses the feedback to observe the pattern in the actual pictures and try to give meanings to the drawings and practice creating drawings by copying, practicing, and being passionate (with trail). To apply these behaviours, after checking the skills of the child by evaluating hand pressure *HP*. The current hand pressure is compared with *RHP*; if it is smaller than *RHP*, then the solution will be updated using equation (3), taking into account the skill rate *(SR)* and level rate *(LR)* of the child, which are two random numbers between (0-1) initially and later to be between (0.6-1) if the child does have a relevant hand pressure. Setting *SR* and *LR* to high (0.6-1) indicates that the child does have an accurate level of knowledge and skill rate, though it can be developed by considering the *GR* factor. Another factor used to update the solution and improve its performance is the Golden Ratio *(GR)*. *GR* is the ratio between the two chosen factors of the solution, which are the length and width of a child's drawing (see equation (5)). Each one of these two factors is randomly selected among all of the factors of the problem using (equation (6)).

$$X_{i+1} = GR + SR \cdot * (X_{ilbest} - X_i) + LR \cdot * (X_{igbest} - X_i) \qquad (4)$$
$$X_{iGR} = \frac{X_{iL} + X_{iW}}{X_{iL}} \qquad (5)$$
$$L, W = rand(0, j) \qquad (6)$$

In Eq (4), $X_{ilbest}$ is the child's best drawing, so far, that is the local best solution, and $X_{igbest}$ is the global best solution observed by the children in their environment? Besides, the Golden Ratio (GR) is the ratio between length *(L)* and width *(W)* of the child's drawing.

### 4.1.4 Fourth stage (Creativity)

Each child has the creativity and skills gained through experience and observing the surrounding environment. Creativity is a factor that makes any art piece to be more visually appealing. In this stage, the child is combining information to update the solutions that have a golden ratio or near to the golden ratio. However, the solution does not have a relevant hand pressure that indicates that a child's skills are not much developed yet and needs improvement using the creativity factor as well as the golden ratio. Moreover, any child memories the best learning practices and try to imitate the same process to get better results. For that, a Pattern Memory *(PM)* is created for each solution in the algorithm; the size of the pattern can change according to the problems. However, selecting a random solution among the *PM* array to be used for updating the solutions that are not performing well is one of the techniques to increase the convergence rate of the algorithm, and in real life, it escalates the learning speed of the children. Both *CR* and *PM* are applied in (equation (7)), which is used in updating the current solution and converging toward the optimal solution. The improvement is added by considering the creativity factor, which is a fixed value set by trial and error to be *CR*=0.1. Later, both *SR* and *LR* are set to low (0-0.5) indicates that the child does not have an accurate level of knowledge and skill rate, though it can be developed by considering creativity rate and pattern memory.

$$X_{i+1} = X_{iMP} + CR \cdot * (X_{igbest}) \qquad (7)$$

### 4.1.5 Fifth stage (Pattern memory)

Adding some other details and being more precise, and comparing them to all the best drawings by using previous knowledge and skills. This behaviour is applied in the algorithm by choosing one of the best ten child's best drawings randomly to update the current drawing, which has an accurate golden ratio but the hand pressure is not relevant. The stage mainly focuses on detailed inches of the drawings. The behaviour is applied in the algorithm since it is manifested by the agent's personal best updating mechanism. It is when the solution will be updated if there are better solutions, and this is true for updating the global best solution of the population. This behaviour will also be true while updating the pattern memory with the current best global solution reached so far in each iteration.

### *4.2 Algorithm flow and pseudo code of CDDO*

The steps of the new proposed algorithm and its procedure are as follows shown in Figure 9 and the visualization of the flow chart shown in Figure 10:

1. Initialize a child's drawing population.
2. Evaluate drawings, set personal and group's best drawings.
3. Calculate the golden ratio of each drawing.
4. Create initial pattern memory from the best drawings and choose one of them to be utilized for updating the current drawing.
5. Update the drawings either by using the golden ratio or pattern memory according to the child's hand pressure and the golden ratio.
6. Update the level rate and skill rate of the child.
7. Evaluate the cost values, update Personal, global best, and pattern memory.

*Begin*

*Initialize child's drawing population $X_i$ (i = 1, 2, ..., j)*

*Compute each drawing's fitness, set personal best and global best*

*Calculate the golden ratio of each drawing eq. (5)*

*Create pattern memory array*

*Randomly choose an index of pattern memory*

*While (t < maximum number of iterations)*

*Calculate RHP using eq. (2)*

*Randomly choose hand pressure P1 eq. (3) Length P2, Width P3 eq. (6)*

    *For each drawing*

        *if (hand pressure was low)*

        *Update the drawings using eq. (4)*

            *Set LR and SR to HIGH (0.6-1)*

        *Endif*

        *Elseif (XiGR is near to golden ratio)*

        *Consider the learnt patterns, LR and SR using eq. (7)*

        *Set LR and SR to LOW (0-0.5)*

        *Endif*

        *Evaluate the cost values*

        *Update Personal, global best*

        *Update pattern memory*

        *Store the Best Cost Value*

*End for*

*Increment t*

*End while*

***End***

*Return Global best*

**Figure 9: CDDO algorithm Pseudo-code**

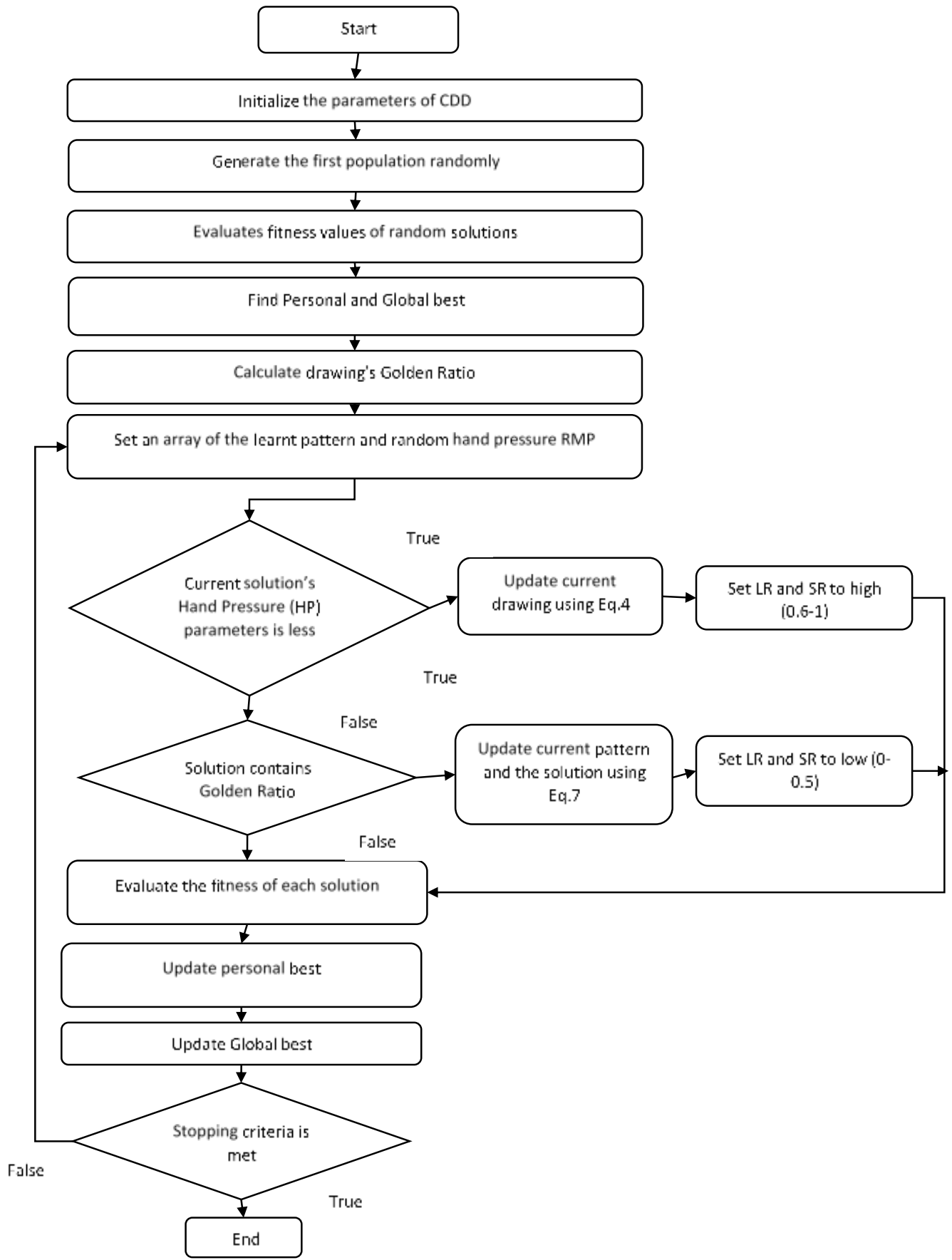

**Figure10: CDDO Algorithm flow chart**

## 5 Testing, Results, and Discussion

In optimization, a standard approach is applied by researchers to confirm a new algorithm's performance as well as to be able to compare their results with the state of the art and classical algorithms. This is a widely used approach aiming at evaluating the optimizers on a large test set, particularly in optimization researches. Nevertheless, it is also important to note that testing the algorithm by these benchmark functions and comparing the results must not be the only factor for evaluating their performance as some of them are very specialized and with no diverse properties. The kind of problems must be targeted, in which the algorithm has a better performance compared to others. In that way, the algorithm's performance can be evaluated more appropriately. For making it possible, test areas must be large enough so that a wide range of problems can be included, such as unimodal, multimodal, and multi-dimensional problems. In this research, the effectiveness of the proposed algorithm has been tested by solving 19 classical benchmark functions utilized in the optimization literature, further explained in the below section that defines a collection of 19 unconstrained optimization test problems used to validate the performance of optimization algorithms.

### *5.1 Benchmark Functions*

The benchmark functions are mathematical optimization problems used to evaluate the effectiveness of the optimization algorithm in finding the best solution. The important strategies used by these benchmark functions, which are shifting, rotating, expanding, and combined variants of these benchmark functions that are the most complicated mathematical optimization problems presented in the literature. From F1 to F7 (see Table 1), are Unimodal functions properly for the exploitation of the variants since these functions have only one global optimum and no local optima. Multimodal functions from F8 to F13 (See Table 2) are utilized for examining local optima avoidance. Multimodal functions test exploration capacity that shows the capability of the algorithm to start from local optima and carry on the search in a wide range of parts of the search space. In other words, they evaluate the exploration of the variants for a large number of local optima. The third part of these benchmark functions is fixed dimension functions which are from F14 to F19 (see Table 3).

Table 1: Unimodal benchmark functions.

| Range [-100,100], Fmin =0 $f_1(x) = \sum_{i=1}^{n} x_i^2$ 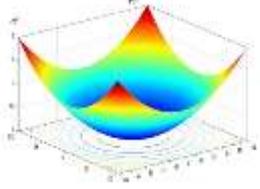 | Range [-10,10], Fmin =0 $f_2(x) = \sum_{i=1}^{n}|x_i| + \prod_{i=1}^{n}|x_i|$ 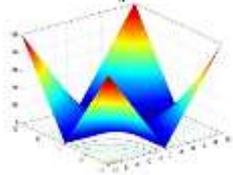 | Range [-100,100], Fmin =0 $f_3(x) = \sum_{i=1}^{n}(\sum_{j-1}^{i} x_j)^2$ 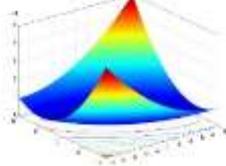 | Range [-100,100], Fmin =0 $f_4(x) = max_i\{|x_i|, 1 \leq |i| \leq n\}$ 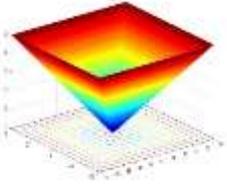 |
|---|---|---|---|
| Range [-30,30], Fmin =0 $f_5(x) = \sum_{i=1}^{n-1}[100(x_{i+1} - x_i^2)^2 + (x_i - 1)^2]$ 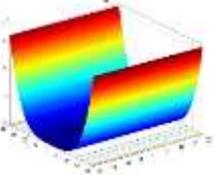 | Range [-100,100], Fmin =0 $f_6(x) = \sum_{i=1}^{n}([x_i + 0.5])^2$ 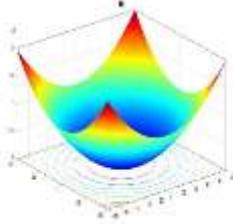 | Range [-1.28,1.28], Fmin =0 $f_7(x) = \sum_{i=1}^{n} i x_i^4 + random[0,1]$ 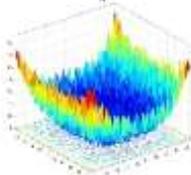 | |

Table 2: Multimodal benchmark functions.

| Range [-500,500], Fmin =-418.9829x5 $f_8(x) = \sum_{i=1}^{n} -x_i \sin(\sqrt{|x_i|})$ 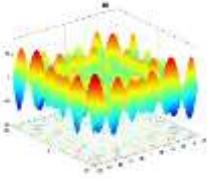 | Range [-5.12,5.12], Fmin =0 $f_9(x) = \sum_{i=1}^{n}[x_i^2 - 10\cos(2\pi x_i) + 10]$ 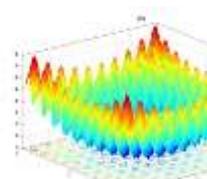 | Range [-32,3] Fmin =0 $f_{10}(x) = -20\exp\left(-0.2\sqrt{\frac{1}{n}\sum_{i=1}^{n} x_i^2}\right) - \exp\left(\frac{1}{n}\sum_{i=1}^{n}\cos 2\pi x_i + 20 + e\right)$ 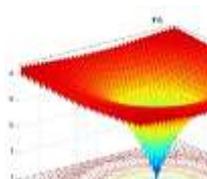 |
|---|---|---|
| Range [-50,50], Fmin =0 $f_{12}(x) = \frac{\pi}{n}\{10\sin(\pi y_1) + \sum_{i=1}^{n-1}(y_i - 1)^2[1 + 10\sin^2(\pi y_{i+1})] + (y_n - 1)^2\} + \sum_{i=1}^{n} u(x_i, 10, 100, 4)$ $y_i = 1 + \frac{x_i+1}{4}$ $u(x_i, a, k, m) = \begin{cases} k(x_i - a)^m & x_i > a \\ 0 & -a < x_i < a \\ k(-x_i - a)^m & x_i < -a \end{cases}$ 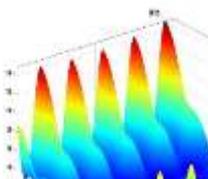 | Range [-50,50] , Fmin =0 $f_{13}(x) = 0.1\{sin^2(3\pi x_1) + \sum_{i=1}^{n}(x_i - 1)^2[1 + sin^2(3\pi x_i + 1)] + x_n - 1)^2[1 + sin^2(2\pi x_n)]\} + \sum_{i=1}^{n} u(x_i, 5, 100, 4)$ 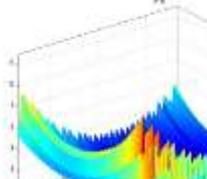 | |

Table 3: Fixed-dimension multimodal benchmark functions.

| | |
|---|---|
| Range [-65,65], Fmin =1, Dim=2 $f_{14}(x) = \left(\frac{1}{500} + \sum_{j=1}^{25} \frac{1}{j+\sum_{i=1}^{2}(x_i - a_{ij})^6}\right)^{-1}$ 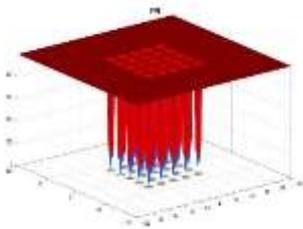 | Range [-5,5], Fmin =0.00030, Dim=4 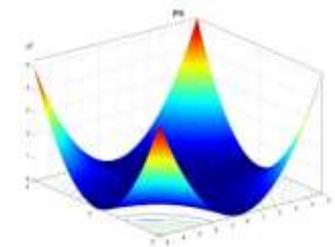 $f_{15}(x) = \sum_{i=1}^{11}\left[a_i - \frac{x_1(b_i^2 + b_i x_2)}{b_i^2 + b_i x_3 + x_4}\right]^2$ |
| Range [-5,5], Fmin =-1.0316, Dim=2 $f_{16}(x) = 4x_1^2 - 2.1 x_1^4 + \frac{1}{3}x_1^6 - x_1 x_2 - 4x_2^2 + 4x_2^4$ 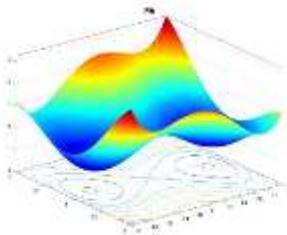 | Range[-5,5], Fmin =0.3982, Dim=2 $f_{17}(x) = \left(x_2 - \frac{5.1}{4\pi^2}x_1^2 + \frac{5}{\pi}x_1 - 6\right)^2 + 10\left(1 - \frac{1}{8\pi}\right)\cos x_1 + 10$ 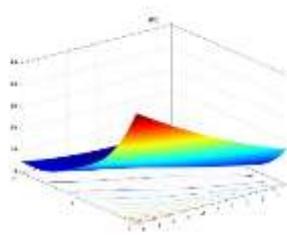 |
| Range [-2,2], Fmin =3, Dim=2 $f_{18}(x) = [1 + (x_1 + x_2 + 1)^2 (19 - 14 x_1 + 3 x_1^2 - 14 x_2 + 6 x_1 x_2 + 3 x_2^2)]$ $\times [30 + (2x_1 - 3 x_2)^2$ $\times (18 - 32 x_1 + 12 x_1^2 + 48 x_2 - 36 x_1 x_2 + 27 x_2^2)]$ 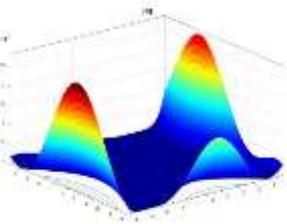 | Range [1,3], Fmin =-3.86, Dim=3 $f_{19}(x) = -\sum_{i=1}^{4} c_i \exp\left(-\sum_{j=1}^{3} a_{ij}(x_j - p_{ij})^2\right)$ 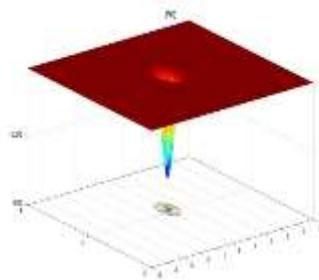 |

## 5.2 Numerical Experiments

The CDDO was coded and evaluated against Whale Optimization Algorithm (WOA), Particle Swarm Optimization (PSO), Gravitational Search Algorithm (GSA), Differential Evolution (DE), and Fast Evolutionary Programming (FEP) algorithms in MATLAB R2016a and implemented on Intel(R), Pentium-Intel Core, i5 Processor(2.90ghz), 430 M, 500GB HDD, 4GB Ram.

## 5.3 Parameter Setting

In algorithms, CDDO, WOA, PSO, GSA, DE, and FEP's parameters were set as follows:
1. Number of search agents (candidate) = 30;
2. Maximum number of iterations (generations) = 500;
3. *CR*=0.1.

## 5.4 Results and Evaluations

In this section, the results obtained by the CDDO are shown, and it is compared with the results of the following algorithms WOA, DE, PSO, GSA, and FEP, published [35], [36]. The benchmark functions were implemented in MATLAB, and parameters of the CDDO algorithm were set randomly according to the range of each function. The average fitness (AVG) and the corresponding standard deviation (StdDev) result obtained of each benchmark function, along with the StdDev and AVG of the CDDO algorithm, are summarised in Tables 4 and 5. The comparison between CDDO and other algorithms are summarised in Table 6 below.

In the below subsections CDDO result is demonstrated, studied, and compared with other algorithms based on its ability in exploitation, exploration, and local minima avoidance.

1) Exploitation:

Along with the proposed algorithm, other considered algorithms were compared and run 30 times on each benchmark function to test and evaluate not only the exploration capacity of the algorithm in finding new solutions but also exploiting toward the best solution while avoiding local minima. Unimodal benchmark functions shown in Table-1 starting from F1 to F7 are used to measure the ability of the CDDO to lead the search agents towards the best solution. The numerical results for F1-F7 demonstrate that CDDO outperforms other algorithms for most of the test cases of Unimodal benchmark functions. Table 4 shows the average and standard deviation of results for CDDO compared to WOA, PSO, GSA, DE, FEP algorithms. It could be declared CDDO algorithm has remarkable performance as it was the most efficient among all the considered algorithms, which are WOA, PSO, GSA, DE, FEP for the benchmark functions in Unimodal, including F1, F2, F3, F7. At the same time, CDDO has very competitive results in F4, F5, and F6, as shown in Tables 4 and 5. This makes CDDO have a high convergence speed towards optimum and exploit it accurately on Unimodal functions.

2) Exploration:

In contrast to Unimodal functions, multimodal functions contain many local optima, which increase according to the number of design variables. According to the aforementioned statement, using multimodal functions show the capacity of the algorithm in exploring the search space and finding new solutions. CDDO has used more than one method to find new search areas that are relevant by using both Creativity and Pattern Memory. They both contribute effectively in the process of randomizing the search space and generating solutions that have a new direction of the search space. Mainly this is achieved by introducing random parameters to the process. Presented results of F8-F13 in Table 4 shows that CDDOA has a competitive result compared to other algorithms considered in this research, specifically F8-F10, which performs better than most of the considered algorithms in this research, namely DE, PSO, GSA algorithms. As shown in (Table 6), CDDO has different ranges of levels in multimodal functions; however, it still has a noticeable result compared to other algorithms.

3) Local minima avoidance:

Local minima avoidance is among the challenging tasks since only by an appropriate balance between exploration and exploitation, local optima avoidance can be assured to some extent. To test the capability of the algorithm in avoiding local minima, multimodal Fixed-dimension multimodal benchmark functions are used due to having a massive number of local minimal. The numerical results of the benchmark functions starting from F14 to F19 can demonstrate CDDO is at least better than one of the considered algorithms or more regarding local minima avoidance except for F16- F18, in which the result of the CDDO algorithm has lower performance than the other algorithms. However, It can be seen that CDDO is competitive enough with other state-of-the-art meta-heuristic algorithms as it does have stand-out results, particularly in F14, F17, and F18 (Tables 4, 5). This shows the capability of the algorithm to avoid local minima as they cover promising regions extensively within the design space and exploit the best solution. The approach used to ensure avoiding local minima is the fact that all of the search agents change and abrupt in the early stages of the optimization process and then converge gradually toward the best solution. This approach will later guarantee that the search agents that work cooperatively eventually convergences to a point in a search space.

Table 4: CDDO algorithm result comparison using standard deviation

| Func | CDDO | WOA | PSO | GSA | DE | FEP |
|------|------|-----|-----|-----|----|----|

|     | Std | Std | Std | Std | Std | Std |
|-----|-----|-----|-----|-----|-----|-----|
| F1  | 3.6E-57 | 4.9E-30 | 2.0E-04 | 9.7E-17 | 5.9E-14 | 1.3E-04 |
| F2  | 6.3E-30 | 2.4E-21 | 4.5E-02 | 1.9E-01 | 9.9E-10 | 7.7E-04 |
| F3  | 1.4E-38 | 2.9E-06 | 2.2E+01 | 3.2E+02 | 7.4E-11 | 1.4E-02 |
| F4  | 3.8E-32 | 4.0E-01 | 3.2E-01 | 1.7E+00 | 0.0E+00 | 5.0E-01 |
| F5  | 2.1E+01 | 7.6E-01 | 6.0E+01 | 6.2E+01 | 0.0E+00 | 5.9E+00 |
| F6  | 1.3E+00 | 5.3E-01 | 8.3E-05 | 1.7E-16 | 0.0E+00 | 0.0E+00 |
| F7  | 1.1E-03 | 1.1E-03 | 4.5E-02 | 4.3E-02 | 1.2E-03 | 3.5E-01 |
| F8  | 1.8E+03 | 7.0E+02 | 1.2E+03 | 4.9E+02 | 5.7E+02 | 5.3E+01 |
| F9  | 2.1E+01 | 0.0E+00 | 1.2E+01 | 7.5E+00 | 3.9E+01 | 1.2E-02 |
| F10 | 3.1E-15 | 9.9E+00 | 5.1E-01 | 2.4E-01 | 4.2E-08 | 2.1E-03 |
| F11 | 2.5E-01 | 1.6E-03 | 7.7E-03 | 5.0E+00 | 0.0E+00 | 2.2E-02 |
| F12 | 3.8E-02 | 2.1E-01 | 2.6E-02 | 9.5E-01 | 8.0E-15 | 3.6E-06 |
| F13 | 4.5E-01 | 2.7E-01 | 8.9E-03 | 7.1E+00 | 4.8E-14 | 7.3E-05 |
| F14 | 2.7E-03 | 2.5E+00 | 2.6E+00 | 3.8E+00 | 3.3E-16 | 5.6E-01 |
| F15 | 2.5E-03 | 3.2E-04 | 2.2E-04 | 1.6E-03 | 3.3E-04 | 3.2E-04 |
| F16 | 4.7E-03 | 4.2E-07 | 6.3E-16 | 4.9E-16 | 3.1E-13 | 4.9E-07 |
| F17 | 1.4E-02 | 2.7E-05 | 0.0E+00 | 0.0E+00 | 9.9E-09 | 1.5E-07 |
| F18 | 3.2E-01 | 4.2E-15 | 1.3E-15 | 4.2E-15 | 2.0E-15 | 1.1E-01 |
| F19 | 1.0E-01 | 2.7E-03 | 2.6E-15 | 2.3E-15 | N/A | 1.4E-05 |

Table 5: CDDO algorithm result comparison using average

| Func | CDDO | WOA | PSO | GSA | DE | FEP |
|------|------|-----|-----|-----|-----|------|
|      | Avg  | Avg | Avg | Avg | Avg | Avg |
| F1   | 1.3E-57 | 1.4E-30 | 1.4E-04 | 2.5E-16 | 8.2E-14 | 5.7E-04 |
| F2   | 1.9E-30 | 1.1E-21 | 4.2E-02 | 5.6E-02 | 1.5E-09 | 8.1E-03 |
| F3   | 2.8E-39 | 5.4E-07 | 7.0E+01 | 9.0E+02 | 6.8E-11 | 1.6E-02 |
| F4   | 7.8E-33 | 7.3E-02 | 1.1E+00 | 7.4E+00 | 0.0E+00 | 3.0E-01 |
| F5   | 3.3E+01 | 2.8E+01 | 9.7E+01 | 6.8E+01 | 0.0E+00 | 5.1E+00 |
| F6   | 1.5E+00 | 3.1E+00 | 1.0E-04 | 2.5E-16 | 0.0E+00 | 0.0E+00 |
| F7   | 1.2E-03 | 1.4E-03 | 1.2E-01 | 8.9E-02 | 4.6E-03 | 1.4E-01 |
| F8   | -1.1E+04 | -5.1E+03 | -4.8E+03 | -2.8E+03 | -1.1E+04 | -1.3E+04 |
| F9   | 2.0E+01 | 0.0E+00 | 4.7E+01 | 2.6E+01 | 6.9E+01 | 4.6E-02 |
| F10  | 7.5E-15 | 7.4E+00 | 2.8E-01 | 6.2E-02 | 9.7E-08 | 1.8E-02 |
| F11  | 6.5E-02 | 2.9E-04 | 9.2E-03 | 2.8E+01 | 0.0E+00 | 1.6E-02 |
| F12  | 3.8E-02 | 3.4E-01 | 6.9E-03 | 1.8E+00 | 7.9E-15 | 9.2E-06 |
| F13  | 6.8E-01 | 1.9E+00 | 6.7E-03 | 8.9E+00 | 5.1E-14 | 1.6E-04 |
| F14  | 1.0E+00 | 2.1E+00 | 3.6E+00 | 5.9E+00 | 1.0E+00 | 1.2E+00 |
| F15  | 1.9E-03 | 5.7E-04 | 5.8E-04 | 3.7E-03 | 4.5E-14 | 5.0E-04 |
| F16  | -1.0E+00 | -1.0E+00 | -1.0E+00 | -1.0E+00 | -1.0E+00 | -1.0E+00 |
| F17  | 4.1E-01 | 4.0E-01 | 4.0E-01 | 4.0E-01 | 4.0E-01 | 4.0E-01 |
| F18  | 3.2E+00 | 3.0E+00 | 3.0E+00 | 3.0E+00 | 3.0E+00 | 3.0E+00 |
| F19  | -3.7E+00 | -3.9E+00 | -3.9E+00 | -3.9E+00 | N/A | -3.9E+00 |

Table 6: A comparison study of CDDO Algorithm

| Functions | 1st | 2nd | 3rd | 4th | 5th | 6th | Rank | Subtotal |
|-----------|-----|-----|-----|-----|-----|-----|------|----------|
| F1        | CDDO | WOA | PSO | GSA | FEP | DE | 1 | |
| F2        | CDDO | WOA | DE  | PSO | GSA | FEP | 1 | |

| | | | | | | | |
|---|---|---|---|---|---|---|---|
| F3 | CDDO | FEP | WOA | DE | PSO | GSA | 1 | |
| F4 | WOA | CDDO | DE | PSO | FEP | GSA | 2 | |
| F5 | DE | WOA | CDDO | FEP | GSA | PSO | 3 | |
| F6 | DE,FEP | GSA | PSO | CDDO | WOA | - | 4 | |
| F7 | CDDO | WOA | DE | GSA | PSO | FEP | 1 | 13 |
| F8 | DE | CDDO | FEP | WOA | PSO | GSA | 2 | |
| F9 | WOA | FEP | CDDO | GSA | PSO | DE | 3 | |
| F10 | CDDO | DE | FEP | GSA | PSO | WOA | 1 | |
| F11 | DE | WOA | PSO | FEP | CDDO | GSA | 5 | |
| F12 | DE | FEP | PSO | CDDO | WOA | GSA | 4 | |
| F13 | DE | FEP | PSO | CDDO | WOA | GSA | 4 | 19 |
| F14 | DE | FEP | CDDO | WOA | PSO | GSA | 3 | |
| F15 | DE | FEP | WOA | PSO | CDDO | GSA | 4 | |
| F16 | FEP | PSO,GSA,DE,WOA | CDDO | - | - | - | 3 | |
| F17 | PSO, GSA, DE | FEP | WOA | CDDO | - | - | 4 | |
| F18 | PSO,GSA,DE,WOA | FEP | CDDO | - | - | - | 3 | |
| F19 | PSO | GSA | WOA | FEP | CDDO | - | 5 | 22 |
| | | | | | | Total: | 54 |
| | | | | | | Overall Rank: | 53/19=2.8421 |
| | | | | | | F1-F7: | 13/7=1.8571 |
| | | | | | | F8-F13: | 19/6=3.1667 |
| | | | | | | F14-F19: | 22/6=3.6667 |

## 6 Conclusion

To develop an algorithm, a significant review was carried out for the background information of the works done previously in the context. As an inspiration source of this algorithm, children's behaviour was used in drawing development stages and the development of their cognitive skills. Another concept of this algorithm is the Golden ratio. It is utilized in this research as a mathematical factor due to its unique properties used in many fields such as architecture, art, and design.

To test and validate the results gained after applying the proposed algorithm, some mechanisms were used based on the previously used methods by other researchers. The performance of the CDDO algorithm was evaluated, and the test result indicated CDDO is quite competitive by gaining 2.8 ranks among 19 benchmark functions. Throughout the testing process, the effectiveness of the proposed algorithm was evaluated by solving all of the 19-benchmark functions, and different levels of the ranking were noticed in different types of problems, for example, in Unimodal functions, which are F1-F7, CDDO ranked 1.9 among the other algorithms. In Multimodal benchmark functions that include F8-F13 CDDO ranked 3.2, this shows CDDO is noticeably strong in exploring a new solution. This shows the capability of the algorithm to avoid local minima as they cover promising regions extensively within the design space and exploit the best solution.

**Acknowledgements:** The authors would like to thank the University of Kurdistan Hewler for providing facilities for this research work.
**Funding**: This study was not funded.
**Compliance with Ethical Standards**
**Conflict of Interest:** The authors declare that they have no conflict of interest.

**Ethical Approval:** This article does not contain any studies with human participants or animals performed by any of the authors.

**References**

[1] I. Boussaïd, J. Lepagnot, and P. Siarry, "A survey on optimization metaheuristics," *Inf. Sci.*, vol. 237, pp. 82–117, 2013.

[2] C. Blum and A. Roli, "Metaheuristics in combinatorial optimization: Overview and conceptual comparison," *ACM Comput. Surv. CSUR*, vol. 35, no. 3, pp. 268–308, 2003.

[3] M. Madić, D. Marković, and M. Radovanović, "Comparison of meta-heuristic algorithms for solving machining optimization problems," *Facta Univ.-Ser. Mech. Eng.*, vol. 11, no. 1, pp. 29–44, 2013.

[4] D. M. Hutton, "The quest for artificial intelligence: A history of ideas and achievements," *Kybernetes*, 2011.

[5] P. Agarwal and S. Mehta, "Nature-inspired algorithms: state-of-art, problems and prospects," *Int. J. Comput. Appl.*, vol. 100, no. 14, pp. 14–21, 2014.

[6] L. Abualigah, D. Yousri, M. Abd Elaziz, A. A. Ewees, M. A. Al-qaness, and A. H. Gandomi, "Aquila Optimizer: A novel meta-heuristic optimization Algorithm," *Comput Indus Eng Httpsdoi Org101016j Cie*, 2021.

[7] Y. Zhang, S. Wang, and G. Ji, "A comprehensive survey on particle swarm optimization algorithm and its applications," *Math. Probl. Eng.*, vol. 2015, 2015.

[8] S. Mirjalili, S. M. Mirjalili, and A. Lewis, "Grey wolf optimizer," *Adv. Eng. Softw.*, vol. 69, pp. 46–61, 2014.

[9] A. Abraham, S. Das, and S. Roy, "Swarm intelligence algorithms for data clustering," in *Soft computing for knowledge discovery and data mining*, Springer, 2008, pp. 279–313.

[10] S. P. Adam, S.-A. N. Alexandropoulos, P. M. Pardalos, and M. N. Vrahatis, "No free lunch theorem: A review," *Approx. Optim.*, pp. 57–82, 2019.

[11] L. Amodeo, E.-G. Talbi, and F. Yalaoui, *Recent developments in metaheuristics*. Springer, 2018.

[12] X.-S. Yang, *Nature-inspired optimization algorithms*. Academic Press, 2020.

[13] L. Abualigah and A. Diabat, "Advances in sine cosine algorithm: a comprehensive survey," *Artif. Intell. Rev.*, pp. 1–42, 2021.

[14] M. Kumar and A. J. Kulkarni, "Socio-inspired optimization metaheuristics: a review," *Socio-Cult. Inspired Metaheuristics*, pp. 241–265, 2019.

[15] M. Bhuvaneswari, S. Hariraman, B. Anantharaj, and N. Balaji, "Nature inspired algorithms: A review," *Int. J. Emerg. Technol. Comput. Sci. Electron.*, vol. 12, no. 1, pp. 21–28, 2014.

[16] M. Dixit, N. Upadhyay, and S. Silakari, "An exhaustive survey on nature inspired optimization algorithms," *Int. J. Softw. Eng. Its Appl.*, vol. 9, no. 4, pp. 91–104, 2015.

[17] M. Dorigo and G. Di Caro, "Ant colony optimization: a new meta-heuristic," in *Proceedings of the 1999 congress on evolutionary computation-CEC99 (Cat. No. 99TH8406)*, 1999, vol. 2, pp. 1470–1477.

[18] Z. W. Geem, J. H. Kim, and G. V. Loganathan, "A new heuristic optimization algorithm: harmony search," *simulation*, vol. 76, no. 2, pp. 60–68, 2001.

[19] D. Karaboga, "An idea based on honey bee swarm for numerical optimization," Citeseer, 2005.

[20] I. Fister, I. Fister Jr, X.-S. Yang, and J. Brest, "A comprehensive review of firefly algorithms," *Swarm Evol. Comput.*, vol. 13, pp. 34–46, 2013.

[21] X.-S. Yang, *Nature-inspired metaheuristic algorithms*. Luniver press, 2010.

[22] X.-S. Yang, "Nature-inspired mateheuristic algorithms: success and new challenges," *ArXiv Prepr. ArXiv12116658*, 2012.

[23] A. S. Shamsaldin, T. A. Rashid, R. A. Al-Rashid Agha, N. K. Al-Salihi, and M. Mohammadi, "Donkey and smuggler optimization algorithm: A collaborative working approach to path finding," *J. Comput. Des. Eng.*, vol. 6, no. 4, pp. 562–583, 2019.

[24] J. M. Abdullah and T. Ahmed, "Fitness dependent optimizer: inspired by the bee swarming reproductive process," *IEEE Access*, vol. 7, pp. 43473–43486, 2019.

[25] L. Abualigah, A. Diabat, S. Mirjalili, M. Abd Elaziz, and A. H. Gandomi, "The arithmetic optimization algorithm," *Comput. Methods Appl. Mech. Eng.*, vol. 376, p. 113609, 2021.

[26] U. Goswami and P. Bryant, "Children's cognitive development and learning," 2007.

[27] J. Einarsdottir, S. Dockett, and B. Perry, "Making meaning: Children's perspectives expressed through drawings," *Early Child Dev. Care*, vol. 179, no. 2, pp. 217–232, 2009.

[28] M. Akhtaruzzaman and A. A. Shafie, "Geometrical substantiation of Phi, the golden ratio and the baroque of nature, architecture, design and engineering," *Int. J. Arts*, vol. 1, no. 1, pp. 1–22, 2011.

[29] H. E. Huntley, *The divine proportion*. Courier Corporation, 2012.

[30] A. Fiorenza and G. Vincenzi, "From Fibonacci sequence to the golden ratio," *J. Math.*, vol. 2013, 2013.

[31] J. Hufford, "An overview of the developmental stages in children's drawings," *Marilyn Zurmuehlen Work. Pap. Art Educ.*, vol. 2, no. 1, pp. 2–7, 1983.

[32] T. Akseer, M. G. Lao, and S. Bosacki, "Children's Gendered Drawings of Play Behaviours," *Alta. J. Educ. Res.*, vol. 58, no. 2, pp. 300–305, 2012.

[33] J. Trawick-Smith, *Early childhood development: A multicultural perspective*. Pearson Higher Ed, 2013.

Cite as: Sabat Abdulhameed and Tarik A. Rashid (2021) Child Drawing Development Optimization Algorithm Based on Child's Cognitive Development, Arabian Journal for Science and Engineering. DOI : 10.1007/s13369-021-05928-6
[34] P. Vasant, *Handbook of research on novel soft computing intelligent algorithms: Theory and practical applications*. IGI Global, 2013.
[35] S. Mirjalili and A. Lewis, "The whale optimization algorithm," *Adv. Eng. Softw.*, vol. 95, pp. 51–67, 2016.
[36] L. M. Q. Abualigah, *Feature selection and enhanced krill herd algorithm for text document clustering*. Springer, 2019.